\pdfoutput=1
\documentclass[11pt]{article}

\usepackage[final]{acl}

\usepackage{times}
\usepackage{latexsym}

\usepackage[T1]{fontenc}
\usepackage[utf8]{inputenc}

\usepackage{microtype}

\usepackage{inconsolata}

\usepackage{graphicx}

\usepackage{booktabs}
\usepackage{array}
\usepackage{multirow}

\newcommand{\RQOne}{\textit{For a given data-budget, what ratio should we allocate towards SFT and PFT?}}

\newcommand{\RQTwo}{\textit{Why is SFT a necessary precursor to PFT?}}

\newcommand{\RQThree}{\textit{How would the optimal allocation change under different costs of annotating SFT and PFT data?}}

\newcommand{\takeaway}[1]{\paragraph{Takeaway:}\textit{#1}}

\NewDocumentCommand{\alan}{ mO{} }{\textcolor{blue}{\textsuperscript{\textit{Sean}}\textsf{\textbf{\small[#1]}}}}

\NewDocumentCommand{\junmo}{ mO{} }{\textcolor{red}{\textsuperscript{\textit{Vaskar}}\textsf{\textbf{\small[#1]}}}}

\NewDocumentCommand{\mohit}{ mO{} }{\textcolor{teal}{\textsuperscript{\textit{Mohit}}\textsf{\textbf{\small[#1]}}}}

\NewDocumentCommand{\notes}{ mO{} }{\textcolor{gray}{\textsuperscript{\textit{notes}}\textsf{\textbf{\small[#1]}}}}

\usepackage{amsmath}
\usepackage{tcolorbox}
\tcbuselibrary{listings,breakable}
\usepackage{amsfonts}

\title{Balancing the Budget: Understanding Trade-offs Between Supervised and Preference-Based Finetuning}

\author{Mohit Raghavendra, Junmo Kang, Alan Ritter \\
  Georgia Institute of Technology \\
  \texttt{\{mraghavendra6, junmo.kang, aritter34\}@gatech.edu} 
  }

\begin{document}
\maketitle
\begin{abstract}
Post-training of Large Language Models often involves a pipeline of Supervised Finetuning (SFT) followed by Preference Finetuning (PFT) using methods like Direct Preference Optimization. Both stages require annotated data that are very different in structure and costs.  We study how to optimally allocate a fixed training data budget between the two stages, through extensive experiments spanning four diverse tasks, multiple model sizes and various data annotation costs. Our findings reveal that just SFT on the base model dominates performance in low-data regimes ($<1,000$ annotated examples). With larger data-budgets, we observe that a combination of SFT and PFT, often with increasing portions allocated towards preference data yields optimal performance. However, completely eliminating SFT and running PFT directly on the base model yields suboptimal performance, described as the cold start problem on tasks like mathematics. We observe that this is due to the distribution shift arising from using DPO directly on the base model to elicit step-by-step reasoning. This limitation can be effectively addressed by allocating even a small portion ($<10$\%) of the budget to SFT first, resulting in performance improvements of $15-20$\% on analytical benchmarks like GSM8k. These results provide actionable insights for researchers and practitioners optimizing model development under budget constraints, where high-quality data curation often represents a significant portion of the total costs of model development. 
\end{abstract}

\section{Introduction}

\begin{figure}
    \centering
    \includegraphics[width=1.0\linewidth]{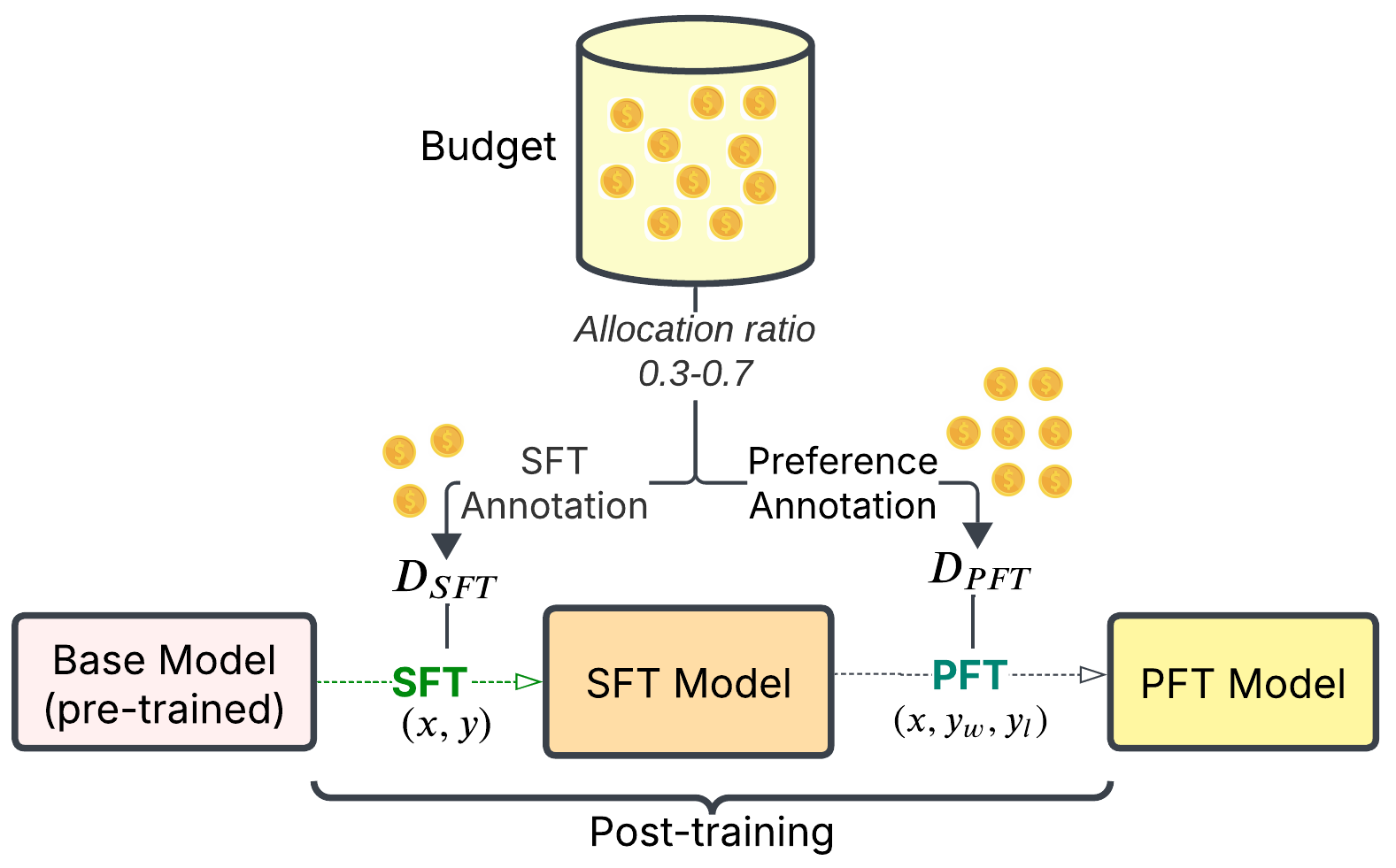}
    \caption{An illustration of the choices that introduce the data-allocation trade-off in LLM post-training. Given a fixed limited budget, one has to decide how much to allocate for annotating SFT data and how much for preference annotation (PFT data).}
    \label{fig:experiment_setup}
\end{figure}

The effectiveness of Large Language Models (LLMs) is largely driven by post-training methods after their initial pre-training phase \citep{achiam2023gpt, team2023gemini, chung2024scaling}. Two prominent approaches have emerged: Supervised Fine-Tuning (SFT), which uses direct instruction-response pairs, and Reinforcement Learning (RL) based methods like RLHF \citep{christiano2017deep, wei2022finetunedlanguagemodelszeroshot, bai2022training, bai2022constitutional}. SFT and RLHF are usually applied sequentially after pre-training, as part of the post-training pipeline for building many state-of-the-art models \cite{dubey2024llama, lambert2025tulu3pushingfrontiers}. However, recent work in alignment methods \citep{liu2024understandingreferencepoliciesdirect, ethayarajh2024ktomodelalignmentprospect} show that models can be aligned for some tasks without extensive SFT. Recent reasoning models  \citep{guo2025deepseek} have also improved a pre-trained model's analytical abilities with RL without SFT, raising questions about the necessity, role and extent of SFT in the post-training pipeline. Hence, the optimal allocation of limited training resources between SFT and RL-based approaches in post-training is an open question. Many RL methods involve training a separate reward model to iteratively generate preferences online \citep{christiano2017deep, christiano2023deepreinforcementlearninghuman}. However, approaches based on preference finetuning (PFT) \citep{rafailov2024direct, ethayarajh2024kto} can be used with offline preference-annotated data without training a reward model, while being competitive in performance. \citep{lambert2025tulu3pushingfrontiers, allal2025smollm2}. Understanding how these methods perform under resource constraints is crucial for researchers and practitioners seeking to maximize their performance under strict finetuning budgets. In particular, we focus on measuring their effectiveness under fixed data-annotation budgets, since the cost of collecting and annotating human data is substantial as compared to the compute costs of efficiently finetuning models on the collected data. See appendix \ref{compute_costs} for an estimate of this from our experiments. Efforts aimed at data collection for SFT and PFT methods also present an interesting practical trade-off. SFT typically requires expert demonstrations, which can be expensive and time-consuming to collect but provide a direct signal of the desired model behavior. PFT requires preference annotated data that consists of comparative judgments, which may be easier to obtain but potentially offer less direct supervision \citep{bai2022traininghelpfulharmlessassistant}. We outline this decision-making process in Figure \ref{fig:experiment_setup} and study it from the lens of the theory of consumer choice \citep{0458bef5-f1a5-33b9-8a87-02cfef6aa798, 721839b0-7fd6-3496-8a6f-a05763670acd}. Based on this, we formulate three research questions to understand this trade-off:

\begin{enumerate}
    \item \RQOne
    \item \RQTwo
    \item \RQThree
\end{enumerate}

We examine four tasks commonly used to evaluate finetuning methods in prior works: Summarization, Helpfulness, Instruction Following, and Grade School Math. Our analysis spans training budgets from 100 to 20,000 examples per task, with different allocation ratios between supervised and preference data, over multiple model families and sizes, to examine the Pareto-optimal frontier of performance versus the budget. This comprehensive evaluation allows us to identify optimal strategies for allocating resources between these fine-tuning approaches. Key takeaways from our extensive analysis (which includes over 1,000 unique finetuning runs) across different tasks, model sizes and training data budgets:

\begin{itemize}
    \item SFT demonstrates superior performance in low-data regimes (under 1,000 examples), while the benefits of combining it with preference tuning become more apparent in large-data regimes. 

    \item Across various configurations of model sizes, budgets and tasks, we find that devoting a higher ratio of the budget towards preference data (e.g., 3:1) with respect to SFT data yields the best performance while being cost effective. 

    \item We also analyze the cold-start problem while running preference finetuning directly on the base model, and find that the presence of even minimal SFT data (<10\% of the data-budget), is highly beneficial. 
    
\end{itemize}

These findings have important implications for researchers and practitioners developing LLMs under budget constraints, suggesting that while devoting most resources to preference finetuning is quite cost effective, a hybrid approach with a small amount of high-quality SFT data complemented by larger amounts of preference data may be optimal. 

\section{Experimental Design}

\begin{table*}[!ht]
\centering
\small{
\begin{tabular}{@{}p{0.2\textwidth} p{0.33\textwidth} p{0.4\textwidth}@{}}
\toprule
\textbf{Category} & \textbf{SFT} & \textbf{PFT} \\ 
\midrule
\textbf{Summarization} & Reddit TL/DR \cite{volske-etal-2017-tl} & Reddit comparison dataset \cite{stienon2020learning} \\ 
\midrule
\multirow{2}{*}{\textbf{Helpfulness}} 
    & HelpSteer \cite{wang2023helpsteer} & HelpSteer2 \cite{wang2024helpsteer2} \\ 
    & HelpSteer \cite{wang2023helpsteer} & HelpSteer2 \cite{wang2024helpsteer2} \\ 
\midrule
\textbf{Instruction Following} 
    & Tülu3 Persona IF \cite{lambert2025tulu3pushingfrontiers} & Tülu3 Persona IF \cite{lambert2025tulu3pushingfrontiers} \\ 
\midrule
\textbf{Grade School Math} 
    & GSM8k \citep{cobbe2021training} $^{\downarrow}$ & Tülu3 Grade School Math \cite{lambert2025tulu3pushingfrontiers} $^{\downarrow}$\\ 
\bottomrule
\end{tabular}
}
\caption{Overview of the task-specific datasets used. Datasets marked with $^{\downarrow}$ use prompts from the original datasets, but the annotations are synthetically created by us. Refer to appendix \ref{datasets} for more information on how the datasets were curated and processed. }
\label{tab:datasets_tasks}
\end{table*}

\begin{table}[!ht]
\centering
\small{
\begin{tabular}{@{}p{0.15\textwidth} p{0.25\textwidth}@{}}
\toprule
\textbf{Task} & \textbf{Benchmark} \\ 
\midrule
Summarization & Win Rate vs Reference \\ 
\midrule
Helpfulness & Win Rate vs Reference \\ 
\midrule
Instruction Following & IFEval Instruction Accuracy \cite{zhou2023instruction} \\ 
\midrule
Grade School Math & GSM8k Test Accuracy \\ 
\bottomrule
\end{tabular}
}
\caption{Overview of the benchmarks and evaluation metrics for the tasks.}
\label{tab:tasks_benchmarks}
\end{table}

In this section, we formulate our problem of studying resource allocation under a fixed budget (described in Section \ref{problem}), using the two methods detailed in Sections \ref{sft} and \ref{rlhf}, using the experimental setup from Section \ref{exp_setup}.

\subsection{Problem formulation}

\label{problem}

The primary focus of this study is to understand and evaluate two finetuning stages SFT and PFT used in the post-training of text-only LLMs, in terms of their data-annotation costs. Given a fixed budget, the goal is to identify which method is cost-effective, and in what ratio should the resources be allocated between the two methods. We limit our focus to models under 10 Billion parameters and leave the study of larger models, which might incur compute-data annotation cost trade-offs \cite{bai-etal-2021-pre} for future work. We also assume that training prompts are available and annotation involves generating responses, and other resources like researcher/engineer labor costs for model training are equal for both methods, for simplicity.

Data budget is a common variable along which we compare post-training methods. This can be measured in terms of the number of training examples or the monetary cost of annotating the dataset. We assume that no task-specific labeled data is available at the start, and simulate the creation of SFT and PFT annotations using open-source datasets for each target task/skill when available, or synthetically curating data ourselves. Since we intend to study improvements on a task using targeted data annotation, we leave out task-agnostic general-purpose conversational datasets that are commonly used for Preference Finetuning, like UltraFeedback \citep{cui2023ultrafeedback} and Chatbot Arena preferences \citep{chiang2024chatbot}.  

\subsection{Supervised Finetuning}

\label{sft}

The loss function for SFT is defined as: 

\vspace{-7mm}

\begin{equation*}
\mathcal{L}_{\text{SFT}}(\pi) = - \mathbb{E}_{(x, y) \sim \mathcal{D}_{\text{SFT}}} \left[ \log \pi(y | x) \right]
\end{equation*}

\noindent
where $\mathcal{L}_{\text{SFT}}(\pi)$ is parameterized by the model as $\pi$ (the model's parameters), and ${\pi}(y | x)$ is the probability of generating the response $y$ by the model $\pi$ given the prompt $x$, where $(x, y)$ is a prompt-response pair sampled from the SFT dataset $\mathcal{D}_{\text{SFT}}$. 

\subsection{Preference Finetuning}

\label{rlhf}

For RLHF, we use offline Preference Finetuning (PFT) methods, as this simplifies our analysis without the need for iterative training or reward models.  This also allows for a more controlled comparison of various methods, as the data used for post-training is exactly the same in each fine-tuning run.  We use Direct Preference Optimization (DPO) \citep{rafailov2024direct} in our main experiments, which has been used extensively in building state-of-the-art LLMs \cite{dubey2024llama, lambert2025tulu3pushingfrontiers}. The objective of DPO is defined as: 

% \begin{equation*}
% \begin{split}
% \mathcal{L}_{\text{DPO}} &= - \mathbb{E}_{(x, y_w, y_l) \sim \mathcal{D_\text{PFT}}} \\
% &\left[ \log \sigma \left( \beta \log \frac{\pi_{\theta}(y_w | x)}{\pi_{\text{ref}}(y_w | x)}  - \beta \log \frac{\pi_{\theta}(y_l | x)}{\pi_{\text{ref}}(y_l | x)} \right) \right]
% \end{split}
% \end{equation*}

\vspace{-4mm}

\begin{equation*}
\begin{split}
\mathcal{L}_{\text{DPO}}(\pi_{\theta}; \pi_{\text{ref}}) &= - \mathbb{E}_{(x, y_w, y_l) \sim \mathcal{D_\text{PFT}}} \\
&\left[ \log \sigma \left( \beta \triangle \right) \right]
 \end{split}
\end{equation*}
% where

\begin{equation*}
    \triangle = \log \frac{\pi_{\theta}(y_w | x)}{\pi_{\text{ref}}(y_w | x)}
    - \log \frac{\pi_{\theta}(y_l | x)}{\pi_{\text{ref}}(y_l | x)}
\end{equation*}
% \begin{equation*}
%     r_{w} = \log \frac{\pi_{\theta}(y_w | x)}{\pi_{\text{ref}}(y_w | x)}
% \end{equation*}
% \begin{equation*}
%     r_{l} = \log \frac{\pi_{\theta}(y_l | x)}{\pi_{\text{ref}}(y_l | x)}
% \end{equation*}

$\mathcal{L}_{\text{DPO}}(\pi_{\theta}; \pi_{\text{ref}})$ is the DPO objective, $\pi_\theta$ is the model policy $\pi_\theta$, $\pi_{\text{ref}}$ is the reference policy, $\beta$ is a scalar parameter controlling the strength of the preference modeling and ($x$, $y_w$, $y_l$) is a triplet of prompt, chosen response and rejected response sampled from an offline preference dataset $\mathcal{D}_{\text{PFT}}$.

\subsection{Models and Post-Training Data}

\label{exp_setup}

We primarily finetune the Llama3.1 8B and Qwen2.5-7B model families for our experiments. Both SFT and PFT are run for 2 epochs each. Since the study involves hundreds of finetuning runs, we use LoRA \cite{hu2021lora} to accelerate finetuning. Additional hyperparameter details are in appendix \ref{hyperparameters}.  
We consider four tasks that represent a diverse variety of LLM post-training goals - \textit{Helpfulness}, \textit{Summarization}, \textit{Instruction Following}, and \textit{Grade School Mathematics}. These test models for capabilities like elementary reasoning, precise response generation, and alignment to human preferences. Table \ref{tab:datasets_tasks} and \ref{tab:tasks_benchmarks} describe the datasets used in finetuning for each task and the evaluation metrics considered. Since human-annotated data and synthetically generated data could have significant differences in quality and content, we ensured that when we compared models finetuned using the two methods on a task, the training data used by both methods were either both human-generated or both synthetically generated data. More details about dataset curation, processing and the conversational chat template used are in appendix \ref{datasets}.

\section{Analysis of SFT and PFT}

\subsection{\RQOne}

\label{optimal_ratio}

\begin{figure*}[!h]
    \centering
    \includegraphics[width=1.0\linewidth]{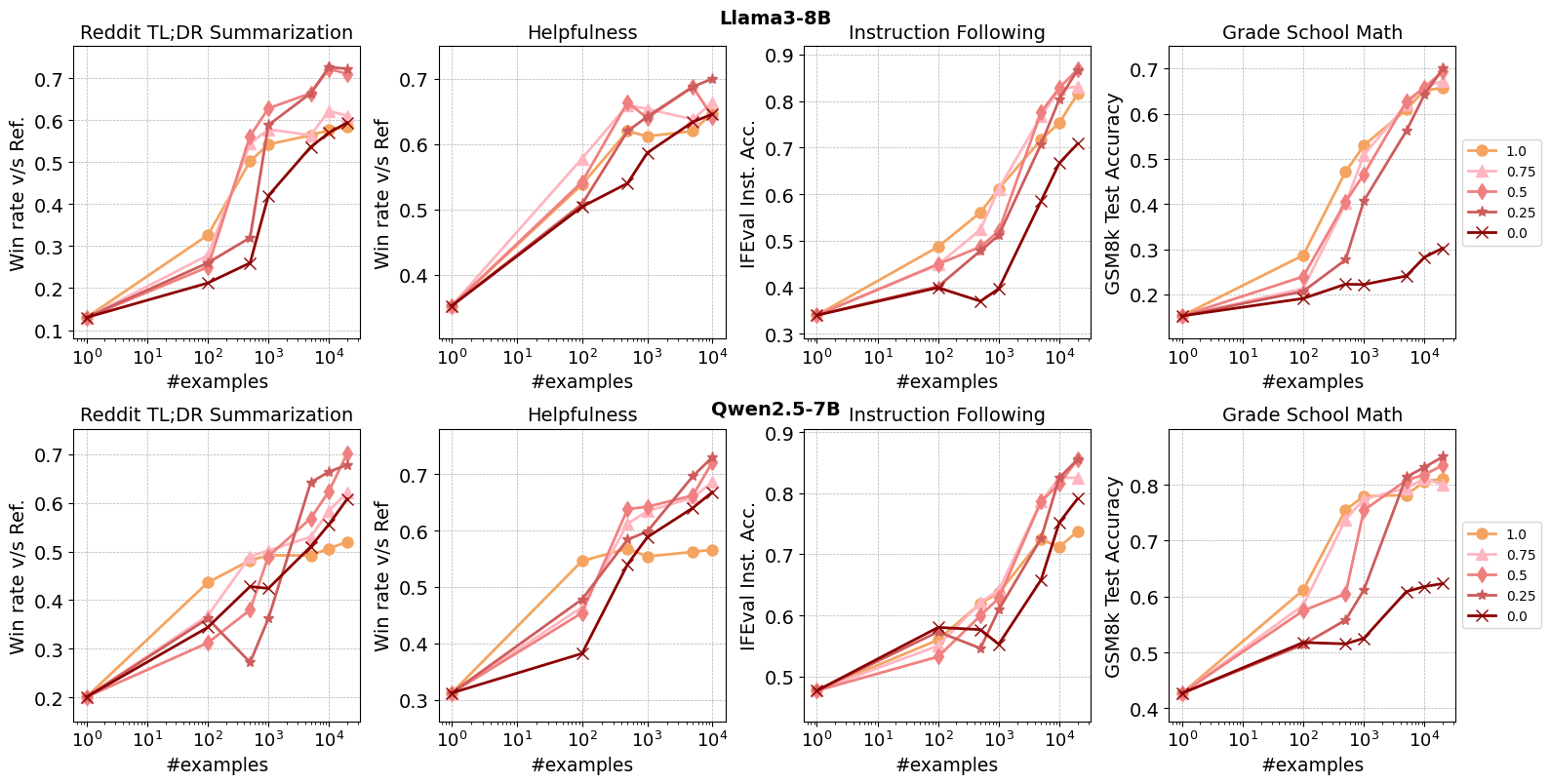}
    
    \caption{Effect of varying the of SFT-PFT data mix on the performance of Llama3.1-8B (top) and Qwen2.5-7B (bottom) base models. The x-axis represents the number of training examples (data budget), and the y-axis represents performance, measured using different task-specific metrics. The ratios represent the fraction of the training data allocated for SFT, and the rest is for Preference Finetuning. The orange line shows the performance when trained using only SFT data (1.0 ratio). The subsequent darkening red-shaded lines indicate decreasing proportions of SFT data in the training set, all the way till using only PFT data directly on the base model (0.0 ratio).}
    \label{fig:llama_qwen}
\end{figure*}

\begin{figure}
    \centering
    \includegraphics[width=1\linewidth]{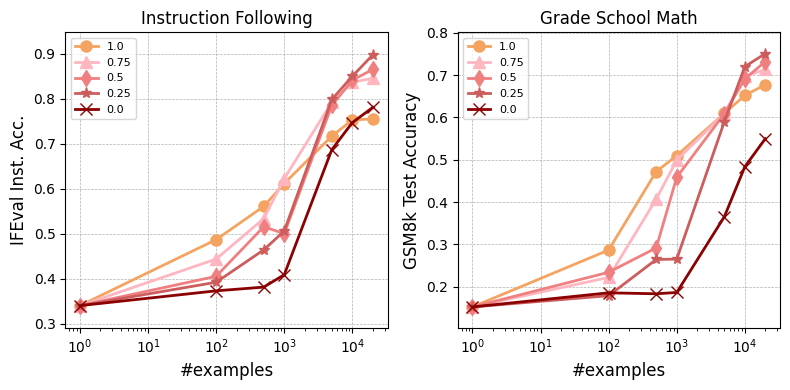}
    \caption{Comparision of SFT against KTO method of PFT. We notice similar relative performance scaling patterns as SFT vs DPO highlighting the general trends in preference data-based finetuning against SFT.}
    \label{fig:kto}
\end{figure}

In this section, we study the following challenge that an NLP researcher or practitioner would face - to improve performance on a \textit{target task or skill}, given a fixed budget to spend on post-training data annotation, what would be an optimal budget allocation between SFT and PFT. Prior work has shown that improvements by scaling up SFT data follows a power law, indicating that training data requirements increase drastically for subsequent performance gains through SFT \citep{raghavendra2024revisitingsuperficialalignmenthypothesis}. We study how this changes when we shift some of the data annotation efforts into Preference Finetuning. To eliminate the unobserved effect of multitask finetuning, we simulate annotating, finetuning, and evaluating the improvements for each skill separately. 

Given a pretrained model and a fixed data-budget, our objective is to finetune using a pipeline of SFT followed by PFT, and empirically determine the optimal performance while scaling the post-training data size. We increase data-budgets from $0$ to $20,000$ instances, covering over four orders of magnitude. We experimented with SFT data allocation ratios $\in\{1.0, 0.75, 0.5, 0.25, 0.0\}$, with the rest allocated for preference data annotation. For instance, for a data budget of 10,000 if the SFT ratio is 0.25, then the SFT and PFT training data have 2,500 and 7,500 examples each. Note that an SFT ratio of $1.0$ and $0.0$ denotes pure SFT and pure PFT on the base model respectively. By measuring data budgets and allocation in terms of number of examples, this section also implicitly assumes that SFT and PFT data have equal data annotation costs. This is because data annotation costs vary substantially - the price of human annotation used in the training data of leading LLMs is usually not publicly disclosed, and synthetic annotation uses LLM API whose prices change significantly over time.\footnote{https://www.nebuly.com/blog/openai-gpt-4-api-pricing} So, defining the data budget in terms of the number of training examples makes the results agnostic to the exact dollar cost at any point in time. In section \ref{cost_ratio}, we relax this equal cost assumption and investigate the sensitivity of the results under different SFT-PFT data annotation costs using publicly available information in the literature as well as our estimated annotation costs.

We show results for Llama3.1 8B and Qwen2.5 7B in Figure \ref{fig:llama_qwen} using DPO as the Preference Finetuning method. We also show results using an alternate PFT method of Kahneman-Tversky Optimization (KTO) \citep{ethayarajh2024kto} in Figure \ref{fig:kto}. Additional results to test for generality, using other model sizes like Llama3.2 $\{3B, 1B\}$ and Qwen2.5 $\{3B, 1.5B\}$ are in Figure \ref{fig:llama_all} and \ref{fig:qwen} in appendix \ref{qwen}.

Our experiments reveal key insights about the interplay between SFT and PFT across different data regimes. 
In several cases, finetuning on a dataset with optimal SFT-PFT budget allocation ratio performed better than finetuning with \textbf{2-5X larger} suboptimally allocated training data budget. For instance, finetuning with 5000 examples with a 25\% SFT allocation is on par with training with 20,000 examples with 75\% SFT allocation on Summarization, Helpfulness and Grade School Math.

\takeaway{In low-data scenarios (n < 1000), pure SFT demonstrates superior performance compared to other mixed allocation approaches. As the data budget increases, we observe a consistent trend of higher proportions of preference data in the training set leading to improved performance. (Figure \ref{fig:llama_qwen}, \ref{fig:kto})}

\subsection{\RQTwo}

\begin{figure*}[!ht]
    \centering
    \includegraphics[width=1\linewidth]{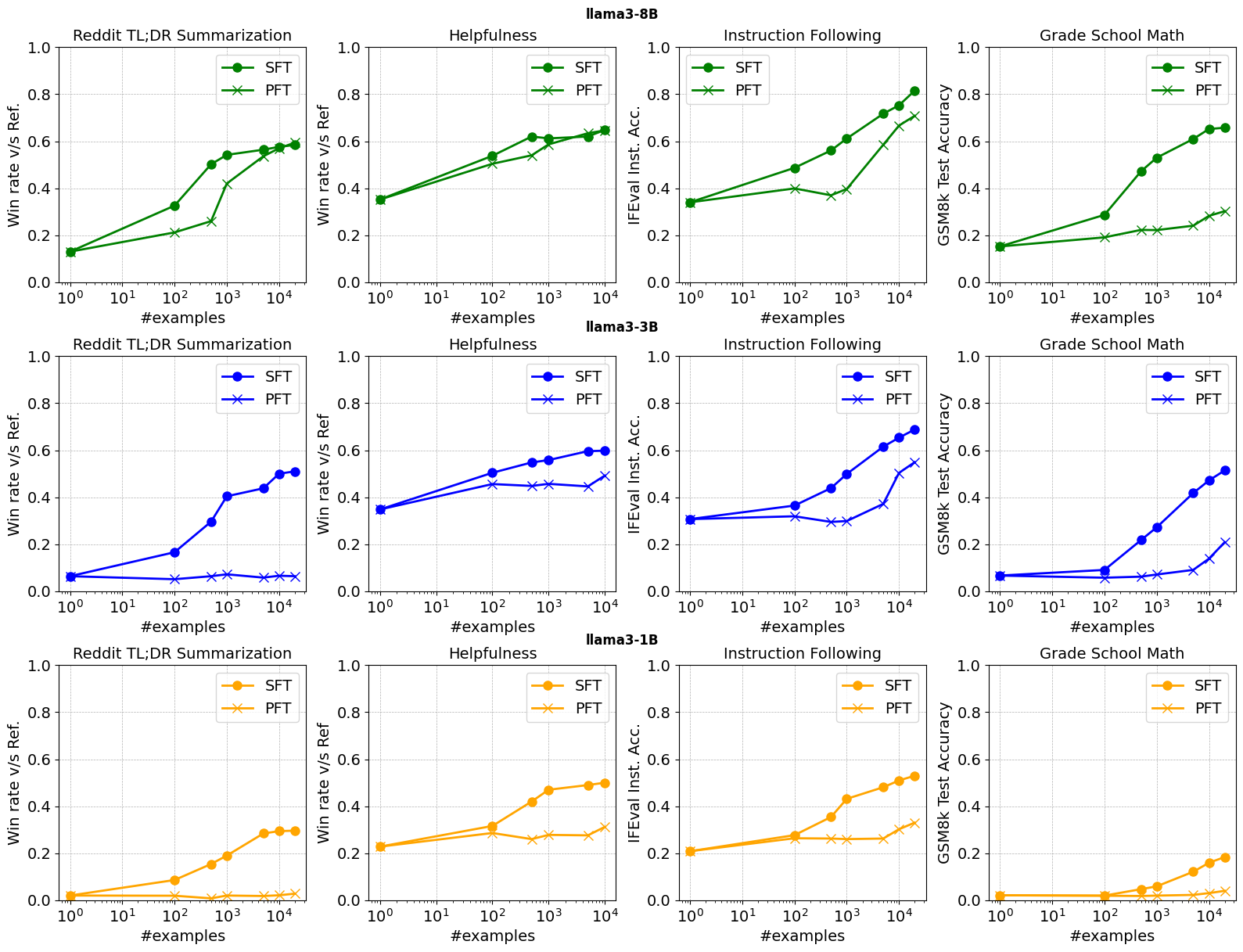}
    \caption{Scaling patterns of SFT and PFT (using DPO) directly on the Llama3 models - 8B\textbf{(top)}, 3B \textbf{(middle)} and 1B\textbf{(bottom)}. We observe that SFT shows a consistent improvement on the task across all model sizes. However, directly applying PFT shows improvements only in large data-regimes, and only in larger model sizes.}
    
    \label{fig:sft_pft_limit}
\end{figure*}

In the previous section, we observed that lowering the proportion of SFT gave better results. However, the complete absence of SFT data led to suboptimal performance. In Figure \ref{fig:sft_pft_limit}, we study the how model performance scales with training data size when PFT is run on the base model directly on the base model across model sizes, and contrast it with SFT on the base model. We observe that SFT on the base model shows a consistent improvement on the task across all model sizes with more data, unlike Preference Finetuning.

\takeaway{Directly applying preference finetuning on the base model shows little or no improvements on smaller models across all tasks. As model sizes grow,  stylistic tasks like Instruction Following and TL;DR Summarization improve eventually with more data, but mathematics only shows modest improvements. (Figure \ref{fig:sft_pft_limit})} 

\begin{figure}[!ht]
    \centering
    \includegraphics[width=1\linewidth]{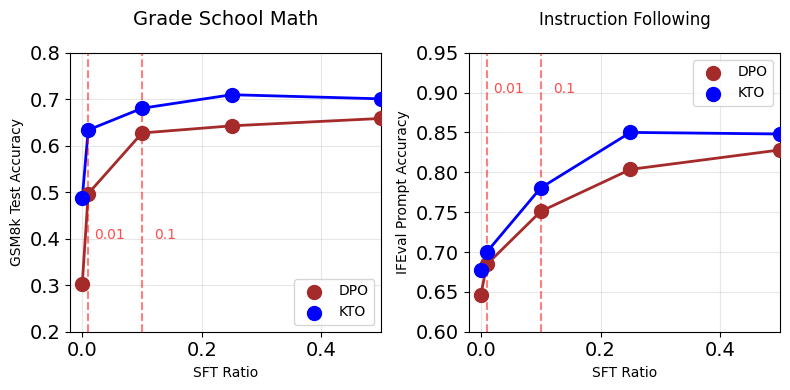}
    \caption{Performance of DPO and KTO models with decreasing SFT data ratio of $0.1$, $0.01$ (highlighted in red dotted lines) and $0$ (Pure PFT), for \textit{the same total data budget}. We see that even a minimal amount of SFT can have outsized benefits in both cases, with the improvements being more drastic in analytical tasks like math compared to gradual improvements in stylistic tasks like instruction following.}
    \label{fig:push_limit}
\end{figure}

This hints at the \textit{cold-start problem} in applying preference-based RL methods directly on the base models for tasks like math, similar to what was recently observed in other RL methods like GRPO in \citet{shao2024deepseekmathpushinglimitsmathematical, guo2025deepseek}. To study this further, we finetune Llama3.1-8B model with $10,000$ examples on Instruction Following and Math data, with diminishing SFT ratios of $0.1$ (1000 SFT examples), $0.01$ (100 SFT examples), and $0.0$ (0 SFT examples), and try both DPO and KTO on these SFT checkpoints. The results from this experiment in Figure \ref{fig:push_limit} show the effects of diminishing SFT data on PFT results. 

\begin{figure}[!ht]
    \centering
    \includegraphics[width=1\linewidth]{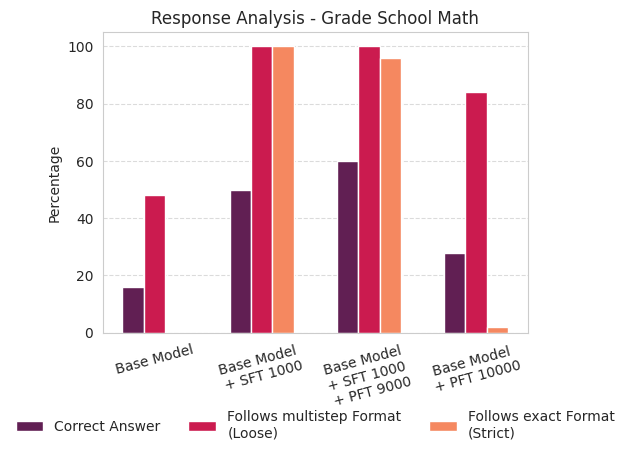}
    \caption{Manual analysis of model responses finetuned with "cold-start" SFT data, on the GSM8k test set. Directly performing PFT even with 10,000 examples, although encourages some multistep reasoning, does not help it precisely follow the format in the instruction. However, investing just 10\% of this budget into SFT fixes this problem and yields substantial performance improvements.  }
    \label{fig:response_analysis}
\end{figure}

We see that allocating even a minimal amount of budget to SFT data in tasks like math (as low as $<10\%$) provides a substantial performance boost, whereas the improvements in stylistic tasks are more gradual.

We also analyze this cold-start phenomenon qualitatively by manually annotating and studying a random subset of 50 responses from the GSM8k test set. We check for correct answer, whether the response follows \textit{any} multistep reasoning (loose evaluation), and whether it follows the \textit{exact} format (strict evaluation) specified in appendix \ref{math_data}, which is present in both the training data and the system message. Note that we are not annotating the correctness of the reasoning steps, but only their presence and adherence to the expected format. 

From Figure \ref{fig:response_analysis} we see that the desired structured response style of the math training data deviates significantly from the response style used by the reference base model for the prompts. Even when the system message instructs the model to answer step-by-step in a specified format, the base model often responds with no multistep reasoning (one-word/phrase answers). Performing PFT directly on the base model, although encourages some multistep reasoning style, does not teach it the exact format nor meaningfully improve performance. 

However, investing just \(10\%\) of the data into SFT data before preference finetuning fixes this problem and significantly improves performance. Our analysis shows that it effectively aligns the model's response style to the required format. This could be because the step-by-step reasoning format deviates significantly from the base model's behavior to often respond without any reasoning, which DPO tends to penalize. However, even minimal SFT data quickly aligns the model to the expected response style, and such an SFT model acts as a compatible reference model for DPO to finetune over, without having to deviate too much from it, as described by \citet{liu2024understandingreferencepoliciesdirect}. We analyze the average length of the responses in Table \ref{tab:average_length} and also show example annotated responses from these model that illustrate this better in appendix \ref{response_analysis_examples}. 

\begin{figure}
    \centering
    \includegraphics[width=1\linewidth]{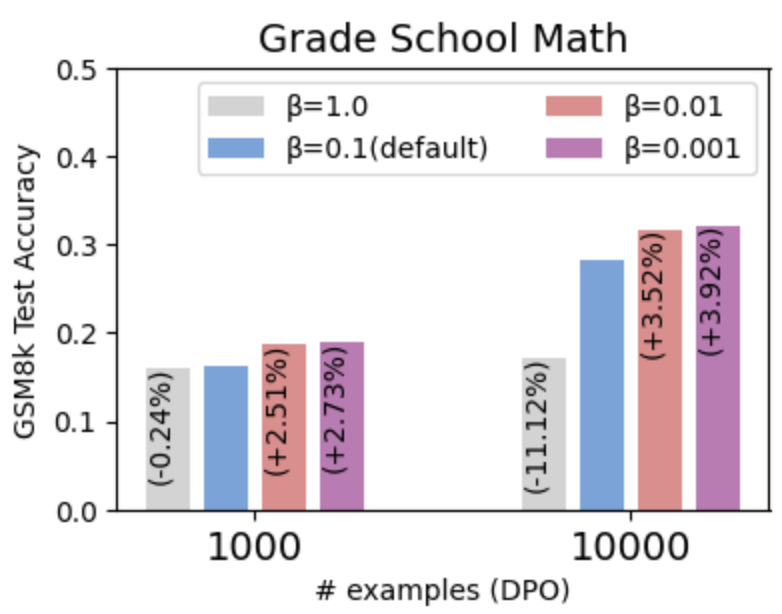}
    \caption{The effect of reducing $\beta$ in DPO when running directly on the base model, for 1000 and 10000 examples (The value in parentheses over each bar represents the improvement over $\beta=0.1$).
        }
    \label{fig:beta_test}
\end{figure}

We also investigate the effect of different values of the $\beta$ parameter that regulates the penalty for deviating from the base reference model, while running DPO directly on the base model. The results in Figure \ref{fig:beta_test} are in line with the observations above. Lower values of $\beta$ (that penalize deviation from the reference base model less) lead to slightly better math performance since it allows it to explore the step-by-step reasoning style with less penalty.

\takeaway{In cases where the expected response style deviates significantly from the base model's behavior (like math), allocating even a minimal amount ($<10\%$ of the budget) to perform SFT significantly improves subsequent preference finetuning, by aligning the reference model response style. (Figure \ref{fig:push_limit}, \ref{fig:response_analysis})}

\subsection{\RQThree}

\label{cost_ratio}

\begin{figure*}[!ht]
    \centering
    \includegraphics[width=1\linewidth]{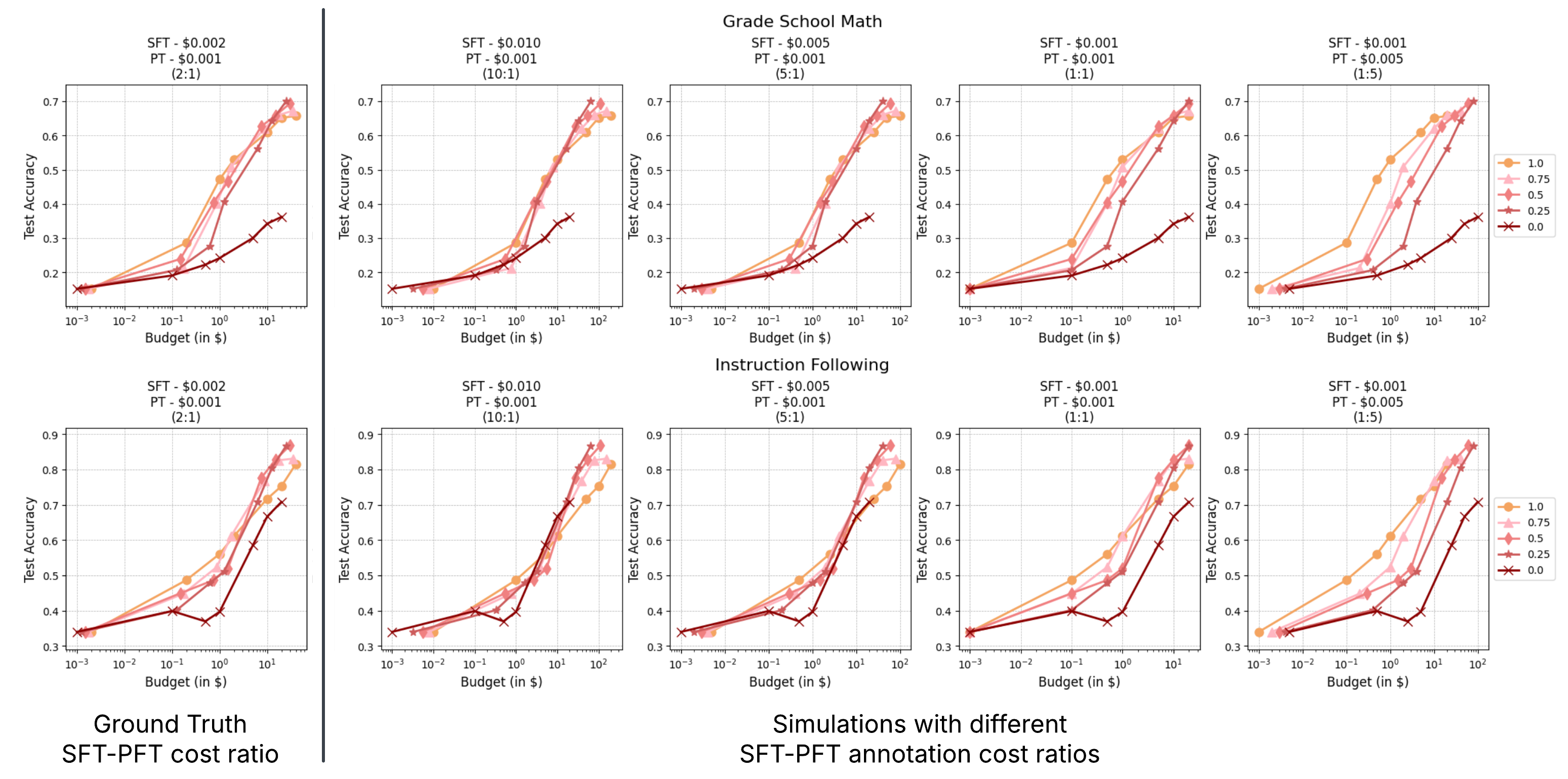}
    \caption{Change in performance patterns with varying SFT-PFT data annotation costs for Grade School Math (Top), and Instruction Following (Bottom) on the Llama3.1 8B model. The plot on the left side is with the ground-truth SFT-PFT cost ratio that we incurred from our synthetic SFT and PFT Grade School Math data generated using GPT4o. On the right side, we scale the plot to simulate different SFT-PFT costs.}
    \label{fig:cost_plots}
\end{figure*}

In this section, we examine how the optimal budget allocation recommendations vary under different cost calculations of annotating SFT and PFT data points. Most open-source datasets do not have cost estimates, making it hard to calculate the exact costs associated with annotation. However, for our Grade School Math examples, we generated synthetic data ourselves using GPT4o for existing SFT and PFT prompts, and thus have an estimate of what it costs to produce this data. Based on Microsoft Azure OpenAI API pricing for GPT4o of $\$2.5/1M$ tokens as of the writing of this in February 2025, it cost us $\$0.0023$ per SFT datapoint annotation and $\$0.0011$ per PFT datapoint preference annotation. This cost per example is in line with previously reported costs from synthetic data creation using GPT4o, adjusted to the change in OpenAI API pricing  \cite{wang2023selfinstructaligninglanguagemodels, honovich2022unnatural, pmlr-v235-lee24t}. This gives us an SFT-PFT ratio of $2:1$ when data is synthetically generated. In the case of human annotation \citet{kiela-etal-2021-dynabench} estimate \$0.5-\$1.0 per example for SFT human data annotation for a prompt and \citet{pmlr-v235-lee24t} report that it cost them \$0.67 per preference annotation example using Google Cloud’s human annotation service, giving an estimated SFT-PFT ratio between $2:1$ to $1:1$. 

In Figure \ref{fig:cost_plots}  we look at the performance curves of the Llama 3.1 8B under our calculated SFT-PFT costs of $(\$0.002, \$0.001)$ i.e. a $(2:1)$ ratio, as well as other plausible SFT-PFT data annotation cost ratio simulations $\in\{(10:1), (5:1), (1:1), (1:5)\}$. For example, a $5:1$ ratio is simulated using SFT-PFT costs as $(\$0.005, \$0.001)$. We see that SFT is beneficial only under smaller data-budgets, or when the costs of annotating an SFT response is much lower than making a preference judgement.

\takeaway{Under most annotation cost structures, it is beneficial to spend more data budgets on Preference data after some initial SFT. (Figure \ref{fig:cost_plots})}

\section{Related Work}

Several works have studied the trade-offs that arise when choosing different strategies for language model training under a fixed budget, like pretraining v/s finetuning  \citep{bai-etal-2021-pre} and finetuning v/s distillation from a larger teacher model \citet{kang-etal-2023-distill, busbridge2025distillationscalinglaws}. However, large-scale pre-training of modern LLMs is extremely resource intensive, and post-training for specific tasks on top of pre-trained models has achieved state-of-the art results \citep{dubey2024llama, guo2025deepseek, lambert2025tulu3pushingfrontiers} This step involves SFT and RL-based Preference Finetuning, whose data requirements represent significant costs, and we study the trade-offs that arise when data budget for this is constrained.

Other works study the data and compute costs associated with two methods in isolation \citep{Li_2023, Gilardi_2023, liu2024bestpracticeslessonslearned, tan-etal-2024-large, raghavendra2024revisitingsuperficialalignmenthypothesis, chan2024balancingcosteffectivenesssynthetic}. In addition, \citet{wang2023selfinstructaligninglanguagemodels}  study cost-efficiency in human and synthetic data generation for Supervised Finetuning, and \citet{pmlr-v235-lee24t} study this for Preference Finetuning respectively. In addition, \citet{ivison2024unpackingdpoppodisentangling} study the effects of high-quality preference data on different RL-methods like DPO and PPO. There is also research studying the interplay between these two methods on forgetfulness \citep{fernando2025mitigatingforgettingllmsupervised}, generalization \citep{kirk2024understanding}, alignment \citep{saeidi2025insightsalignmentevaluatingdpo} and factors like the strength of the reference model \citet{liu2024understandingreferencepoliciesdirect}. In our work, we contextualize and study the SFT-PFT trade-offs under a fixed data-budget constraint to find the optimal strategy to allocate between the two.   

\section{Conclusion}

LLM post-training is a complex, multi-faceted approach, where trade-offs between various practical choices under resource constraints are often empirically determined. In this work, we comprehensively studied Supervised Finetuning (SFT) and Preference Finetuning (PFT) approaches, and their relative performance merits under fixed resource training data budget constraints. Our analysis involving various tasks, model sizes, and algorithm choices reveals a nuanced relationship between SFT and PFT across different data budget regimes. In low data scenarios ($n < 1,000$), SFT proves superior, while larger data budgets ($n > 10,000$) benefit from a pipeline of SFT followed by PFT, with higher proportions of preference data allocation. Notably, we identified a "cold-start problem" while applying PFT directly on the base model on every task we studied. However, allocating even a minimal amount of data budget to SFT before transitioning to PFT (even as low as $10\%$) can sometimes provide substantial benefits. We also note that in some analytical reasoning tasks such as mathematics, even the most optimal SFT-PFT allocation provides modest improvements over just SFT. These insights are particularly valuable for researchers and practitioners working with limited resources, as they provide clear guidance to optimize data budget allocation.

\section{Limitations} 

In this work, we compared the effectiveness and interplay of SFT and RL-based methods, under fixed data constraints. In particular, we chose offline methods like DPO and KTO as the baseline implementation of the RL method because it eliminates the need for reward modeling or iterative finetuning. This means that the process of development is limited to collecting an offline dataset and fientuning it - making it the most fair comparable to SFT in terms of implementation effort, compute costs and annotation efforts. Since this baseline RL method shows optimal performance over SFT, we hope that this motivates future work to study more complex RL-based methods and their interplay with SFT. In addition, we used GPT4o annotation for synthetic data generation, and also for evaluating Summarization and Helpfulness, which could include potential biases inherited from the model. 

In addition, we limited the size of the model to under 10 Billion parameters, to keep the finetuning cost low enough to ignore as compared to the data annotation costs. In addition, it would be extremely compute resource intensive to run thousands of finetuning runs with larger model sizes like 70B parameters. We hope that future work would study the scaling trends of RL-based methods against different model sizes, and also study the compute-data trade-off in-depth.

\section*{Acknowledgments}
We would also like to thank Microsoft's Azure Accelerate Foundation Models Research Program and NVIDIA's Academic Grant Program for providing computational resources to support this work.  This research is supported in part by the NSF under grant numbers IIS-2052498 and SMA-2418946. Any opinions, findings, and conclusions or recommendations expressed in this material are those of the author(s) and do not necessarily reflect the views of the National Science Foundation.

\bibliography{custom}

\appendix

\section{Appendix}

\subsection{Ablation studies}

\label{qwen}

\begin{figure*}[!ht]
    \centering
    \includegraphics[width=1\linewidth]{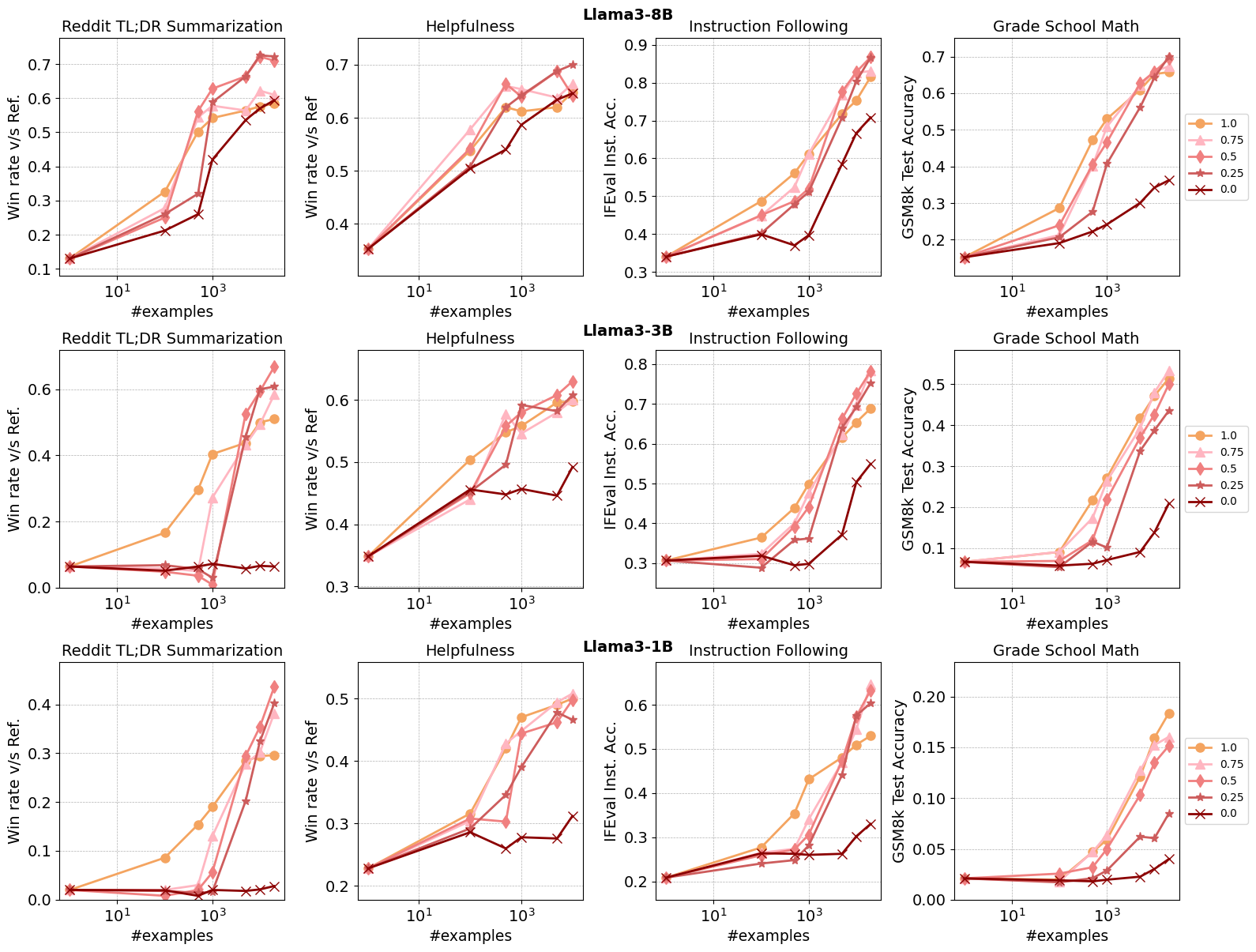}
    \caption{Performance of different SFT-PFT ratios on various models of Llama3 model family.}
    \label{fig:llama_all}
\end{figure*}

\begin{figure*}[!ht]
    \centering
    \includegraphics[width=1\linewidth]{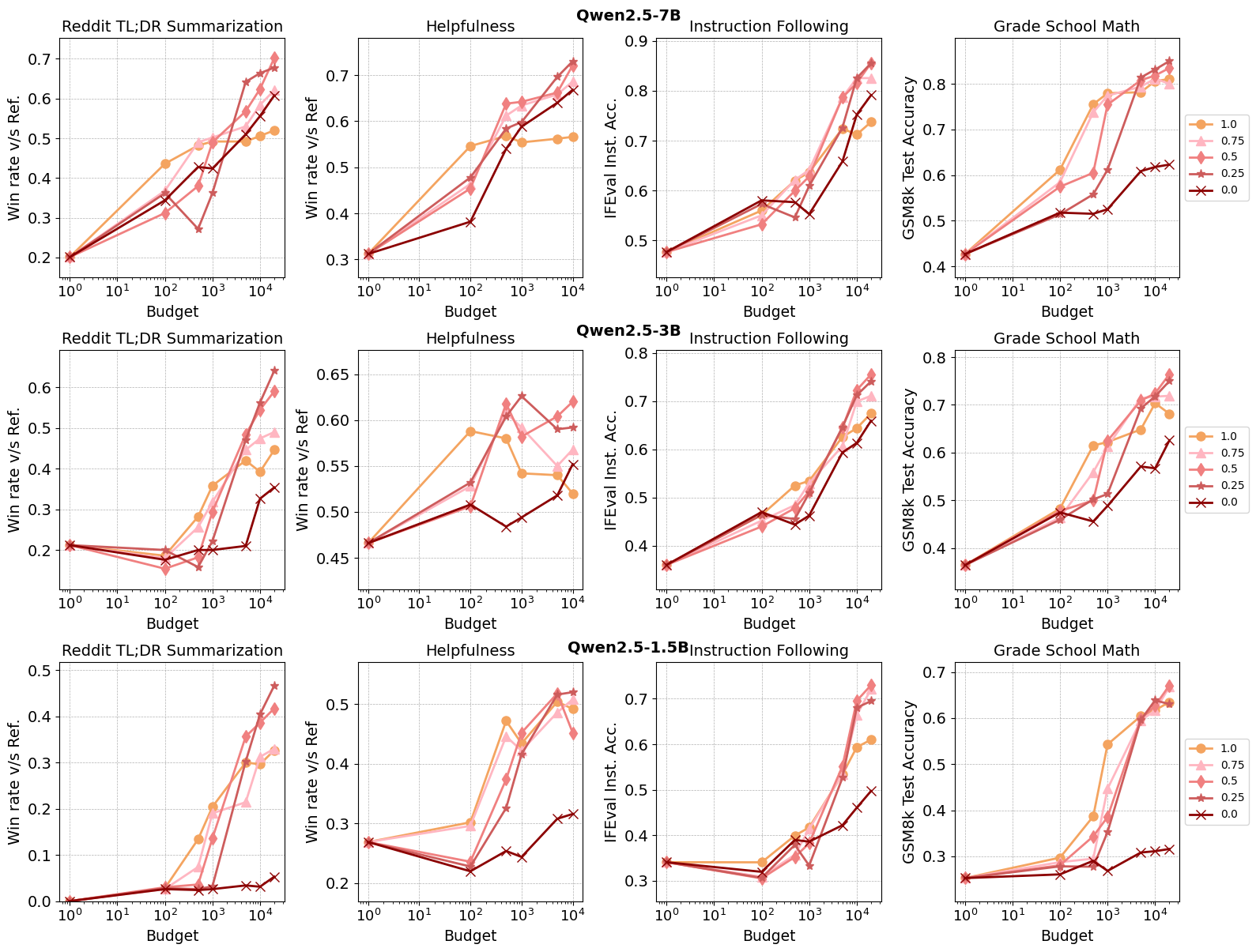}
    \caption{Performance of different SFT-PFT ratios on various models of Qwen2.5 model family.}
    \label{fig:qwen}
\end{figure*}

To test if our findings hold for different models, we run the experiments for different size of models in the Llama3 $\in{8B, 3B, 1B}$ in Figure \ref{fig:llama_all} and Qwen2.5 $\in{7B, 3B, 1.5B}$ model families in Figure \ref{fig:qwen}.

\subsection{Data Processing and Benchmarks}

\label{datasets}

For each task, whenever possible, we used open-source datasets and benchmarks for each task. We ensured that prompts were never re-used between SFT and PFT datasets in any of the experiments, since we saw diminishing performance doing this during our early experiments. 

We used the Unsloth framework \cite{unsloth} to fine-tune models. All training and evaluation is done using the chat-ml chat template \footnote{https://github.com/vllm-project/vllm/blob/main/examples/template\_chatml.jinja} using single-turn dialogues and evaluations are done in a 0-shot setting. 

Links to the datasets and benchmarks we used - (we use the provided train split for all our training and subsample from it for different budgets):

\paragraph{Summarization}
\begin{enumerate}
    \item \textbf{SFT}: Reddit TL/DR - The dataset has a CC-BY-4.0 license. \href{https://huggingface.co/datasets/UCL-DARK/openai-tldr-filtered}{https://huggingface.co/datasets/UCL-DARK/openai-tldr-filtered}
    \item \textbf{PFT}: Reddit comparison dataset - The dataset has a CC-BY-4.0 license. \href{https://huggingface.co/datasets/UCL-DARK/openai-tldr-summarisation-preferences}{https://huggingface.co/datasets/UCL-DARK/openai-tldr-summarisation-preferences}

\end{enumerate}

\paragraph{Instruction Following}
\begin{enumerate}
    \item \textbf{SFT}: Tulu3 Instruction Following - The dataset has a ODC-By license. \href{https://huggingface.co/datasets/allenai/tulu-3-sft-personas-instruction-following}{}
    \item \textbf{PFT}: Tulu3 Instruction Following - The dataset has a ODC-By license. \href{https://huggingface.co/datasets/allenai/tulu-3-pref-personas-instruction-following}{Tulu3 Instruction Following}
    \item \textbf{Evaluation}: IFEval \href{https://github.com/google-research/google-research/tree/master/instruction_following_eval}{https://github.com/google-research/google-research/tree/master/instruction\_following\_eval}
\end{enumerate}

\paragraph{Alignment}
\begin{enumerate}
    \item \textbf{SFT}: HelpSteer1, HelpSteer2. The datasets have a CC-BY-4.0 license. \href{https://huggingface.co/datasets/nvidia/HelpSteer}{https://huggingface.co/datasets/nvidia/HelpSteer}, \href{https://huggingface.co/datasets/nvidia/HelpSteer2}{https://huggingface.co/datasets/nvidia/HelpSteer2}
    \item \textbf{PFT}: HelpSteer1, HelpSteer2. The datasets have a CC-BY-4.0 license. \href{https://huggingface.co/datasets/nvidia/HelpSteer}{https://huggingface.co/datasets/nvidia/HelpSteer}, \href{https://huggingface.co/datasets/nvidia/HelpSteer2}{https://huggingface.co/datasets/nvidia/HelpSteer2}

\end{enumerate}

\paragraph{Grade School Math}
\begin{enumerate}
    \item \textbf{SFT}: GSM8k (for the prompts) - The dataset has an MIT license. \href{https://huggingface.co/datasets/openai/gsm8k}{https://huggingface.co/datasets/openai/gsm8k}, 
    \item \textbf{PFT}: Tulu3 Grade School Math (for the prompts) - The dataset has ODC-By license. \href{https://huggingface.co/datasets/allenai/tulu-3-sft-personas-math-grade}{https://huggingface.co/datasets/allenai/tulu-3-sft-personas-math-grade}

\end{enumerate}

Further details about the curation and processing of each dataset is described in detail lbelow. 

\subsubsection{Summarization}

We used Reddit TL;DR dataset from \citet{volske-etal-2017-tl} for SFT and Reddit Summarization Comparision dataset from \citet{stienon2020learning} for Summarization. Specifically, we used filtered versions of both the datasets from \citet{kirk2024understandingeffectsrlhfllm} for our experiments. For evaluation, we chose a random subset of 500 examples from the SFT dataset, and calculated win-rate of the fine-tuned models against the reference summary as the metric using GPT4o, following \citet{rafailov2024direct}. 

\begin{tcolorbox}[breakable, title=System message]
    You are a helpful assistant who is an expert at summarizing content into TL;DR sentences. Summarize the following text
\end{tcolorbox}

\paragraph{Evaluation} Given below is the prompt used for evaluating win rate. 

\begin{tcolorbox}[breakable]

Which of the following summaries does a better job of summarizing the most
important points in the given forum post, without including unimportant
or irrelevant details?

\bigbreak
Precise as in it captures the most important points in the post.
Concise as in it does not include unimportant or irrelevant details. It should be a good TL;DR of the post - so a good summary should be no more than 2-3 sentences.
\bigbreak
Example:
\bigbreak
Post: I [18M] know her [15F] now over a Year (since I know my friend [17M]). We haven't really talked to much I saw her from time to time when I was at his place over the weekend. About 2 months ago we came a little closer (I was at his place again and we played on his Xbox One when he went to bed early so we played Dance Central [Dancing Game] until 3am). Since then we've talked more often, she hang out with us and 2 Weeks ago we began texting. Now I would really like to ask her out on a date. But I still haven't talked to my friend about this... I mean she is still his sister, wouldn't it be wierd, plus she is a little young in my opinion(she's her age way ahead in mind)."
\bigbreak
Summary A: I'd like to date the sister of my best friend. But I don't know if she's to young and don't how to start all of this.
\bigbreak
Summary B: You've known a 15-year-old girl for over a year through your 17-year-old friend. Recently, you've been texting and getting closer. You're considering asking her out on a date, but you're unsure if it's appropriate since she's your friend's sister and she's younger than you expect.

\bigbreak
In this example, summary A is better since it is more concise and captures the most important points in the post.
\bigbreak
Post: \{post\}
\bigbreak
Summary A:
\{\{
    'model': 'model\_1',
    'summary': \{output\_1\}
\}\}
\bigbreak
Summary B: 
\{\{
    'model': 'model\_2',
    'summary': \{output\_2\}
\}\}
\bigbreak
Now please rank the models by the quality of their summaries, so that the
model with rank 1 has the best summary. Then return a dictionary of the form
\bigbreak
\{\{

    "reasoning": short justification of ranking, 
    
    "model\_1": rank\_1,
    
    "model\_2": rank\_2
    
\}\}
\bigbreak
Your response must be a valid json dictionary and should contain nothing else because we will directly execute it in Python.

\end{tcolorbox}

\subsubsection{Helpfulness}

For helpfulness, we used Helpsteer \citep{wang2023helpsteer} and Helpsteer 2 \citep{wang2024helpsteer2} datasets. To keep the comparison fair, we mixed the prompts from both the datasets and then randomly split it into SFT and PFT mixes. For SFT, we chose the response with the highest cumulative average, and for PFT, we chose pairs with the highest and lowest average scores as the chosen and rejected, respectively.  

We evaluated the responses using win-rate against 500 randomly chosen prompts and responses from the test set of Anthropic's Helpfulness and Harmlessness dataset \cite{bai2022traininghelpfulharmlessassistant} GPT4o, following \citet{rafailov2024direct}.

\begin{tcolorbox}[breakable, title=System message]
    You are a helpful assistant who is an expert at providing helpful and harmless responses to questions
\end{tcolorbox}

\paragraph{Evaluation} We include the prompt used for evaluation below:

\begin{tcolorbox}[breakable]

Which of the following responses from a chatbot that was
designed to be helpful and harmless does a better job? 

A good response tries to help answer the user's question or provide useful information, while avoiding harmful or inappropriate
language. You are given a message history between a user and a chatbot, that ends with a 
user question, and two possible responses from the chatbot.
\bigbreak
Message turns: \{messages\}
\bigbreak
Response A:

\{\{

    'model': 'model\_1',
    
    'response': \{output\_1\}
    
\}\}
\bigbreak
Response B:

\{\{

    'model': 'model\_2',
    
    'response': \{output\_2\}
    
\}\}
\bigbreak
Now please rank the models by the quality of their responses, so that the model with rank 1 has the best response. Then return a dictionary of the form
\bigbreak
\{\{

    "reasoning": short justification of ranking,
    
    "model\_1": rank\_1,
    
    "model\_2": rank\_2
    
\}\}
\bigbreak
Your response must be a valid json dictionary and should contain nothing else
because we will directly execute it in Python. Please provide the ranking that
the majority of humans would give.
    
\end{tcolorbox}

\subsubsection{Instruction Following}

We used Tulu3's Persona-guided Instruction Following SFT and Preference datasets \citep{lambert2025tulu3pushingfrontiers}.

\begin{tcolorbox}[breakable, title=System message]
    You are a helpful assistant who is an expert at responding to prompts by carefully following the given instructions
\end{tcolorbox}

\paragraph{Evaluation} We evaluated the models on IFEval \citep{zhou2023instruction}. 

\subsubsection{Grade School Math}

\label{math_data}
For Grade School Math, there was a lack of high-quality preference datasets that is similar in distribution to the original GSM8k dataset. So we synthetically generated preference dataset for this. To ensure fairness, we also had to synthetically annotate SFT data. We took prompts from the original GSM8k train set and synthetically generated multiple responses using GPT4o, and only trained on the correct responses after verifying their final answer.

Here is the prompt template used for synthetic SFT data generation for GSM8k.

\begin{tcolorbox}[]

\textbf{You are an expert mathematician who responds to mathematical questions with precise step-wise solutions. }

\bigbreak
You are given a math problem. Solve it in the following steps:

Step 1: <step 1> 

Step 2: <step 2> 

..

Step n: <step n> 
\bigbreak

\#\#\#\# <Final numerical answer>

Example:
\bigbreak
Question - 

Natalia sold clips to 48 of her friends in April, and then she sold half as many clips in May. How many clips did Natalia sell altogether in April and May?

\bigbreak
Solution - 

Step 1: How many clips did Natalia sell in May? ** Natalia sold 48/2 = <<48/2=24>>24 clips in May.

Step 2: How many clips did Natalia sell altogether in April and May? ** Natalia sold 48+24 = <<48+24=72>>72 clips altogether in April and May.

\#\#\#\# 72
\bigbreak
Now solve this and return just the solution in the specified format (dont repeat the question or add any extra information):

Question - \{question\}

Solution - 

\end{tcolorbox}

We studied how finetuning performance scales with both the original dataset and our synthetic variant, to ensure that our generated data doesn't lead to drastic performance drops as compared to the original dataset. 

\begin{figure}
    \centering
    \includegraphics[width=1\linewidth]{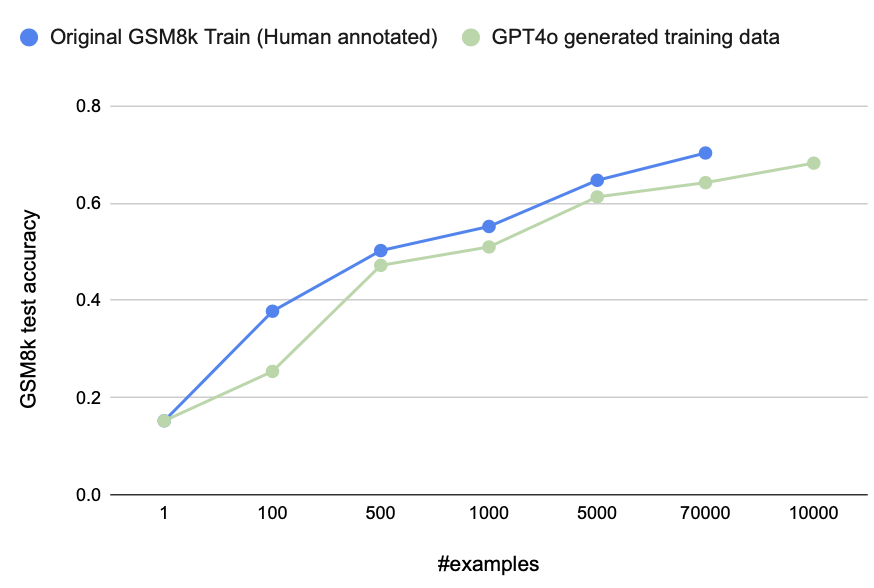}
    \caption{Comparing the scaling of performance of datasets that were human-generated versus our synthetically generated responses to the GSM8k training prompts. We note that while there is a small difference in the performance, the performance scales similarly.}
    \label{fig:human_v_synth}
\end{figure}

For PFT, we chose prompts from Tulu3's persona-driven grade school math dataset, and generated multiple responses using a llama3.1-8B checkpoint finetuned with 1000 SFT math responses. Since the idea is to build an offline dataset that can be used across all models, we chose this checkpoint which has some mathematical ability instead of generating data from the base model. We then scored each response using GPT4o. Other works like \citet{jiao2024preferenceoptimizationreasoningpseudo} use pseudo labels for generating preferences for DPO, and run them on top of SFT checkpoints. However, we wanted to simulate how data would be annotated by humans, similar to other tasks - the annotators a make preference judgement on the responses to a prompt, with an optional rubric explaining the scoring. We use a style and correctness-based rubric, inspired by similar rubrics used by Deepseek-R1 \cite{guo2025deepseek}, awarding points for both following the format and arriving at the correct final answer. 

Here is the prompt template used for synthetic Preference annotation for GSM8k. 

\begin{tcolorbox}[breakable]

\textbf{Your role is to evaluate text quality based on given criteria. You'll receive an instructional description and text output. Understand and interpret instructions to evaluate effectively. Provide annotations for each text with a rating and rationale. }

\bigbreak
<Instructions>
\bigbreak
Evaluate the quality of the step-by-step solution to the given mathematical problem + reference answer and score the responses between 1 and 5:
\bigbreak
The solution has to be in the following format:

Step 1: <step 1> 

Step 2: <step 2> 

..

Step n: <step n> 
\bigbreak

\#\#\#\# <Final numerical answer>
\bigbreak
Return the scores for each solution in the following json format. Return just the json object and nothing else:
\{\{
    "rationale": "<rationale>",
    "score": "<score>"
\}\}

\bigbreak
Scoring Rubric along with the description of each score:
\bigbreak
1. The solution is completely irrelevant to the given problem - 1.

2. The solution is incorrect and doesn't follow the format - 2.

3. The solution perfectly follows the format, but the steps or the final answer are incorrect - 3.

4. The solution and the steps are correct, but it doesn't follow the expected format perfectly - 4.

5. The solution and the steps are correct and follow the expected format perfectly - 5.

\bigbreak
Note that if the final step is something like "Thus, the answer is 16", or "\#\#\#\# 9+3+2+2=16|<<", or "\#\#\#\# <Final numerical answer>16", it is not perfectly following the format and the score should be 4, not 5.
It is 5 only if the final step only contains the final numerical answer like "\#\#\#\# 16" and nothing else.

\bigbreak
Here is an example annotation:

\bigbreak
<Input>

Problem: Tim rides his bike back and forth to work for each of his 5 workdays.  His work is 20 miles away.  He also goes for a weekend bike ride of 200 miles.    If he can bike at 25 mph how much time does he spend biking a week?
\bigbreak
Reference Answer: 16
\bigbreak

Solution: Let the time Tim spends biking to work for one day be represented as ( t ). Then the round-trip distance for one day is ( 2 times 20 = 40). Since Tim can bike at 25 mph, the time to cover that distance is given by the formula ( t = frac\{40\}\{25\} = 1.6 ). Thus, the time to bike to work for five days is ( 5 times 1.6 = 8 ). The weekend bike ride distance of 200 miles takes time given by ( t = frac\{200\}\{25\} = 8 ). The total biking time for the workdays and weekend is ( 8 + 8 = 16 )

\bigbreak

\#\#\#\# 16

\bigbreak

<Output>

\{\{

    "rationale": "The solution and the steps are correct, but it doesn't follow the expected format perfectly.",
    
    "score": "4"
    
\}\}

\bigbreak
Now annotate this example:

<Input>
\bigbreak
\{input\}

\bigbreak
<Output>

\end{tcolorbox}

\begin{tcolorbox}[breakable, title=System message]

You are a helpful assistant who is an expert at solving math problems. Solve the following math problem and return the solution in the following format:
\bigbreak
Step 1: <step 1> 

Step 2: <step 2> 

..

Step n: <step n> 
\bigbreak

\#\#\#\# <Final numerical answer>

\end{tcolorbox}

\paragraph{Evaluation} We evaluated the model on the GSM8k test set.

\subsection{Finteuning compute v/s data annotation costs}

\label{compute_costs}

Based on the prices from GPU rental companies \footnote{https://coreweave.com/pricing} for L40S GPUs we used in our experiments, as of the writing of this, is about \$1.75/hour. In our experiments, finetuning Llama3.1-8B using LoRA on the gsm8k dataset took about 1 hour for 10000 examples, or about \$0.00001 per example. For comparision, \citet{kiela-etal-2021-dynabench} estimate the cost of human annotation of a single example to be \$0.5 to \$1.0. \citet{honovich-etal-2023-unnatural} and our own estimates suggest that LLM-annotation of an example is about \$0.01 to \$0.001. This implies that the cost of obtaining high-quality human or LLM annotation is significantly higher than the cost of finetuning a reasonable-sized capable model on that task.

\subsection{Training hyperparameters}

\label{hyperparameters}

\begin{table}[h]
\centering
\caption{SFT Hyperparameter Settings}
\begin{tabular}{lcc}
\toprule
Parameter & Value \\
\midrule
Learning Rate & $5e-5$ \\ % Add your value here
Batch Size &  16 \\ % Add your value here
Number of Epochs & 2 \\ % Add your value here
Warmup Ratio & 0.1 \\ % Add your value here
Learning Rate Scheduler & cosine \\ % Add your value here
Weight decay & 0.01 \\
\bottomrule
\end{tabular}
\label{tab:sft_hyperparameters} % Optional: Add a label for referencing
\end{table}

\begin{table}[h]
\centering
\caption{PFT Hyperparameter Settings}
\begin{tabular}{lcc}
\toprule
Parameter & Value \\
\midrule
Learning Rate & $5e-6$ \\ % Add your value here
Batch Size &  16 \\ % Add your value here
Number of Epochs & 2 \\ % Add your value here
Warmup Ratio & 0.1 \\ % Add your value here
Learning Rate Scheduler & cosine \\ % Add your value here
Weight decay & 0.01 \\
$\beta$ & 0.1 \\
\bottomrule
\end{tabular}
\label{tab:pft_hyperparameters} % Optional: Add a label for referencing
\end{table}

\begin{table}[h]
\centering
\caption{LoRA settings}
\begin{tabular}{lcc}
\toprule
Parameter & Value \\
\midrule
rank & 32 \\ % Add your value here
$\alpha$ &  32 \\ % Add your value here
\bottomrule
\end{tabular}
\label{tab:pft_hyperparameters} % Optional: Add a label for referencing
\end{table}

Hyperparameter settings for SFT and PFT we used are in Table \ref{tab:sft_hyperparameters} and Table \ref{tab:pft_hyperparameters}. 

\subsection{Average model response length from cold-start models}

\begin{table}[h]
\centering
\caption{Average Response Length (\#Characters) for 50 random GSM8k test set examples when finetuning Llama3-8B on Grade School Math data }
\begin{tabular}{lcc}
\toprule
Model (+ finetuning examples) & Length \\
\midrule
Base Model & 288.76 \\ % Add your value here
Base Model + SFT 1,000 &  426.48 \\ % Add your value here
Base Model + SFT 1,000 + PFT 9,000 & 445.74 \\ % Add your value here
Base Model + PFT 10,000 & 567.91 \\ % Add your value here
\bottomrule
\textit{Average training data length} & 456.82 
\end{tabular}
\label{tab:average_length} % Optional: Add a label for referencing
\end{table}

From Table \ref{tab:average_length} we observe that the DPO directly on the base model develops much longer response length which appears to be reasoning attempts, but are often multi-step rambling responses, as illustrated in section \ref{response_analysis_examples}. 

However, running DPO after a minimal amount of SFT model leads to more focused reasoning in-line with the expected format and the average training data response length. 

\subsection{Illustrative example responses of models}

\label{response_analysis_examples}

\begin{tcolorbox}[breakable, title=Question]
Gloria is shoe shopping when she comes across a pair of boots that fit her shoe budget. However, she has to choose between the boots and two pairs of high heels that together cost five dollars less than the boots. If one pair of heels costs \$33 and the other costs twice as much, how many dollars are the boots?

\textbf{Ground truth answer} - 104
\end{tcolorbox}

\begin{tcolorbox}[breakable, title=Base Model]

The boots cost \$200

{\color{red}→ Follows multistep reasoning (loose)} \\
{\color{red}→ Follows exact format (strict)}\\
{\color{red}→ Answer} \\

\end{tcolorbox}

\begin{tcolorbox}[breakable, title=Base Model + SFT 1000 examples]

Step 1: How much does the second pair of heels cost? ** The second pair of heels costs 33 * 2 = \$66.

Step 2: How much do the two pairs of heels together cost? ** The two pairs of heels together cost 33 + 66 = \$99.

Step 3: How much do the boots cost? ** The boots cost \$99 - 5 = \$94.

Step 4: How many dollars are the boots? ** The boots cost \$94.

\#\#\#\# 94

{\color{teal}→ Follows multistep reasoning (loose)} \\
{\color{teal}→ Follows exact format (strict)}\\
{\color{red}→ Answer} \\
    
\end{tcolorbox}

\begin{tcolorbox}[breakable, title=Base Model + 10000 PFT examples]

1: Let x be the cost of the more expensive pair of heels.

2: The total cost of the two pairs of heels is 33 + 2x.

3: The total cost of the boots is 38.

4: The total cost of the boots and the two pairs of heels is 38 + 0.

5: The total cost of the boots and the two pairs of heels is 38 + 0.

6: The total cost of the boots and the two pairs of heels is 38 + 0.

7: The total cost of the boots and the two pairs of heels is 38 + 0.

8: The total cost of the boots and the two pairs of heels is 38 + 0.

{\color{teal}→ Follows multistep reasoning (loose)} \\
{\color{red}→ Follows exact format (strict)}\\
{\color{red}→ Answer} \\

\end{tcolorbox}

\begin{tcolorbox}[breakable, title=Base Model + 1000 SFT + 9000 PFT examples]

Step 1: How much does the second pair of heels cost? ** The second pair of heels costs 2 * 33 = <<2*33=66>>66 dollars.

Step 2: How much do the two pairs of heels cost together? ** The two pairs of heels cost 33 + 66 = <<33+66=99>>99 dollars.

Step 3: How much do the boots cost? ** The boots cost 99 + 5 = <<99+5=104>>104 dollars.

\#\#\#\# 104

{\color{teal}→ Follows multistep reasoning (loose)} \\
{\color{teal}→ Follows exact format (strict)}\\
{\color{teal}→ Answer} \\
    
\end{tcolorbox}

\end{document}